# Clinical NLP with Attention-Based Deep Learning for Multi-Disease Prediction

Ting Xu
University of Massachusetts Boston
Boston, USA

Xiaoxiao Deng
DePaul University
Chicago, USA

Xiandong Meng
University of California, Davis
Davis, USA

Haifeng Yang
Northeastern University
Boston, USA

Yan Wu*
University of Delaware
Newark, USA

*Abstract-This paper addresses the challenges posed by the unstructured nature and high-dimensional semantic complexity of electronic health record texts. A deep learning method based on attention mechanisms is proposed to achieve unified modeling for information extraction and multi-label disease prediction. The study is conducted on the MIMIC-IV dataset. A Transformer-based architecture is used to perform representation learning over clinical text. Multi-layer self-attention mechanisms are employed to capture key medical entities and their contextual relationships. A Sigmoid-based multi-label classifier is then applied to predict multiple disease labels. The model incorporates a context-aware semantic alignment mechanism, enhancing its representational capacity in typical medical scenarios such as label co-occurrence and sparse information. To comprehensively evaluate model performance, a series of experiments were conducted, including baseline comparisons, hyperparameter sensitivity analysis, data perturbation studies, and noise injection tests. Results demonstrate that the proposed method consistently outperforms representative existing approaches across multiple performance metrics. The model maintains strong generalization under varying data scales, interference levels, and model depth configurations. The framework developed in this study offers an efficient algorithmic foundation for processing real-world clinical texts and presents practical significance for multi-label medical text modeling tasks.*

*Keywords: Electronic medical records; attention mechanism; discrimination algorithm; semantic modeling*

## I. Introduction

With the rapid development of modern medical information systems, Electronic Health Records (EHRs) have become a critical resource for clinical research and intelligent decision-making. In complex medical scenarios such as Intensive Care Units (ICUs), EHR data records not only basic patient information, vital signs, diagnoses, and medications, but also exhibits strong temporal and semantic characteristics. Traditional rule-based or shallow models can no longer meet the demand for high-quality and interpretable medical knowledge extraction. Therefore, exploring more efficient and intelligent information processing mechanisms is essential for advancing precision medicine and personalized care[1,2].

In recent years, deep learning methods, especially those based on pre-trained language models, have shown significant advantages in information extraction tasks. The development of Transformer-based architectures has provided powerful tools for mining latent semantics from unstructured medical texts[3]. Medical language in EHRs is highly domain-specific, context-dependent, and expressed in diverse ways. Models must understand not only semantics but also the structural relationships between medical entities. Against this backdrop, building deep language models capable of capturing the linguistic patterns in medical records and integrating extraction and prediction becomes a key direction in medical AI research.

Medical diagnosis exhibits a clear multi-label nature. Patients often suffer from multiple diseases at once, and EHR texts frequently contain multiple diagnoses. Traditional single-label models show clear limitations under such conditions. They fail to capture the co-occurrence patterns of complex diseases. It is thus necessary to introduce model architectures capable of handling multi-label settings. Integrating domain knowledge further enhances their ability to identify semantic features of disease combinations. This need has driven attention to self-attention mechanisms, enabling the model to capture global dependencies across disease labels and to perceive co-occurrence structures more effectively[4].

In medical AI applications, information extraction supports not only the structuring of static data but also dynamic diagnosis prediction and risk assessment. By modeling diagnostic cues embedded in EHR texts, researchers can help clinicians identify high-risk conditions at earlier stages. This facilitates proactive interventions and resource optimization. In high-risk scenarios like ICUs, improving the identification of disease progression over time and enhancing prediction accuracy directly impacts clinical decisions and patient outcomes. Therefore, multi-label prediction models based on attention mechanisms are both academically challenging and practically valuable.

Focusing on multi-label disease prediction and semantic information extraction in EHRs, it is essential to build deep models centered on Transformer architectures. These models must capture layered semantics and label dependencies. This research direction not only broadens the application scope of natural language processing in healthcare but also provides algorithmic foundations and technical support for improving the efficiency and quality of medical services. It holds significant theoretical value and practical importance.

## II. RELATED WORK AND FOUNDATION

Deep learning has significantly enhanced the capabilities of medical text analysis, particularly for extracting structured insights from the unstructured and semantically dense nature of electronic health records (EHRs). Wang's study on time-aware and multi-source feature fusion introduced an advanced Transformer-based model for integrating diverse data sources in medical text, which aligns closely with the representational demands of clinical narratives [5]. This is further extended by Sun et al., who employed LongFormer architectures for efficient summarization of lengthy EHR texts, addressing challenges in processing extended and context-heavy clinical documents [6].

In tackling multi-label prediction and entity extraction, Wang et al. combined pre-trained language models with few-shot learning strategies to enhance medical entity recognition in sparse-data environments, a key requirement for robust information extraction systems [7]. Zheng et al. proposed a structured gradient guidance approach for adapting large language models to new domains with limited data, which supports the adaptability required in clinical text applications [8]. Complementary to this, Guo et al. presented a perception-guided structural framework to optimize large language model design, enhancing both representation and interpretability [9].

Transformer-based architectures have been increasingly applied to cross-domain and complex clinical tasks. Zhang et al. focused on multi-task learning with unified instruction encoding and gradient coordination, enabling improved performance across various NLP tasks relevant to healthcare applications [10]. Xing contributed a novel prompting strategy for analogical reasoning in pre-trained models, offering potential in adapting models to nuanced clinical relationships through few-shot reasoning [11].

The integration of semantic modeling and structural learning in health informatics is further evidenced by Tayefi et al., who explored analytical frameworks beyond structured data formats, highlighting the need for flexible semantic architectures [12]. Lavin et al. utilized longitudinal EHR data to study disease patterns across demographics, emphasizing real-world applicability of medical text mining techniques [13]. Similarly, Suri et al. evaluated the representativeness of EHR-based cohort studies, advocating for reliable and unbiased patient modeling [14]. Studies focused on anomaly detection and predictive modeling provide additional methodological support. Xin and Pan developed a structure-aware diffusion mechanism for unsupervised anomaly detection in structured datasets, reflecting the importance of structure in medical prediction tasks [15]. Ma applied GANs and temporal autoencoders for anomaly detection in dynamic environments, techniques that can be adapted to time-sensitive clinical settings [16]. Hybrid models and fusion frameworks also play a key role. Yan et al. explored hierarchical feature fusion for robust detection in vision tasks, with potential cross-domain applications in fine-grained medical classification [17]. Fang designed a predictive latency model based on AI-augmented structural modeling, contributing to response optimization strategies in system-critical scenarios [18]. Additionally, Yan et al. applied neural networks to survival prediction across cancer types, demonstrating the transferability of deep architectures to medical risk stratification tasks [19].

Meta-learning and model scalability are tackled in Tang's meta-learning framework, providing a basis for elastic and scalable model deployment across services—an idea that resonates with multi-domain healthcare systems [20]. The foundation for intelligent optimization and system responsiveness is further explored in Sun et al.'s DQN-based approach to dynamic cache management, reflecting reinforcement strategies applicable in intelligent medical systems [21]. These studies collectively underscore the theoretical and practical value of attention-driven and semantically aligned architectures for multi-label classification and structured information extraction in EHRs. They support the methodological foundation of employing deep Transformer-based models enriched with multi-source semantics, adaptive representation, and contextual reasoning to address real-world challenges in clinical informatics.

## III. METHOD

This study builds a disease prediction model based on the Transformer architecture, using electronic medical record text as the primary input, with the goal of achieving collaborative modeling for multi-label disease recognition and semantic information extraction. The process begins by standardizing the clinical text from the original dataset, ensuring consistency in format and content. Following this, the input text is transformed into dense vector representations through a pre-trained medical language model, which captures domain-specific semantics and contextual dependencies. These embeddings are then passed through the Transformer-based encoder to model complex relationships within the text. The overall model design, including the embedding, encoding, and prediction components, is illustrated in Figure 1.

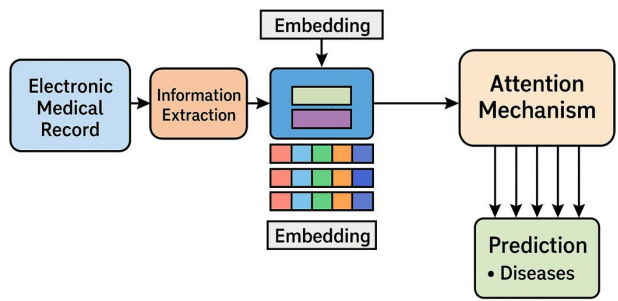

Figure 1. The main model architecture of this article

Suppose the input medical record text sequence is $X = \{x_1, x_2, ..., x_n\}$, and its corresponding word vector is represented by $E = \{e_1, e_2, ..., e_n\}$, where $e_i = R^d$ represents the embedding vector of the i-th word. The embedding vector is then input into the Transformer encoder layer, and the multi-head self-attention mechanism is used to capture the semantic association between words.

During the encoding process, the attention mechanism is expressed by the following formula:

$$Attention(Q, K, V) = \text{softmax}(\frac{QK^T}{\sqrt{d_k}})V \quad (1)$$

Among them, Q, K, and V represent query, key, and value matrices respectively, and A is a scaling factor to alleviate the gradient vanishing problem. The multi-head mechanism calculates multiple attention subspaces in parallel, further enhancing the model's ability to model complex semantic relationships. The output of each head is represented by $head_i = Attention(Q_i, K_i, V_i)$, and the final output is:

$$MultiHead(Q, K, V) = Concat(head_1, ..., head_h)W^O \quad (2)$$

Where A is the learnable projection matrix. The context representation of the model output is mapped to the label space through a fully connected layer for multi-label classification. Let the label set $Y = \{y_1, y_2, ..., y_m\}$ and the model prediction probability be:

$$y'_j = \sigma(W_j^T h + b_j) \quad (3)$$

Where $h$ represents the aggregate representation vector of the text, $W_j$ and $b_j$ are the parameters of the jth label, and $\sigma$ represents the Sigmoid activation function, which ensures that the predicted value of each label falls in the [0,1] interval.

To optimize the multi-label prediction performance, the model is trained using a multi-label loss function based on cross-entropy. The overall loss function is defined as:

$$L = -\frac{1}{m}\sum_{j=1}^{m}[y_j \log(y'_j) + (1-y_j)\log(1-y'_j)] \quad (4)$$

This loss function effectively retains the potential dependencies between different diseases while modeling labels independently, in conjunction with the attention mechanism. The entire training process uses gradient descent to optimize parameters to minimize the loss and improve the generalization ability of the model in multi-label disease recognition tasks.

## IV. EXPERIMENTAL RESULTS

### A. Dataset

This study uses MIMIC-IV (Medical Information Mart for Intensive Care IV) as the primary dataset. The data consists of real ICU electronic health records and includes approximately 500,000 hospital admissions. Collected by a major medical institution in the United States, the dataset contains both structured and unstructured data. It covers multiple dimensions such as patient demographics, diagnoses, medication records, laboratory tests, and vital signs. MIMIC-IV is one of the most widely used open-access databases in medical AI research.

For this task, we mainly use two modules from MIMIC-IV: Clinical Notes and Diagnosis Codes. The Clinical Notes module contains various types of free-text records, including discharge summaries, admission notes, and nursing progress notes. These provide rich semantic context, which helps the language model learn latent medical entities and logical structures. The Diagnosis Codes module offers disease annotations in the ICD (International Classification of Diseases) format. These serve as standardized supervision signals for the multi-label prediction task.

To ensure data quality and reproducibility, we applied several preprocessing steps to the raw texts. These include deduplication, de-identification, code normalization, and sentence segmentation. Only samples with appropriate length and complete labels were retained. The analysis was limited to adult patients. The openness and standardization of the dataset provide a stable foundation for model training and evaluation. This also supports the scalability and comparability of related research.

### B. Experimental Results

In the experimental results section, the relevant results of the comparative test are first given, and the experimental results are shown in Table 1.

Table 1. Comparative experimental results

| Method | Accuracy | Precision | Recall |
|---|---|---|---|
| LSTM[22] | 68.4 | 65.1 | 62.7 |
| CNN[23] | 69.2 | 66.4 | 64.1 |
| BILSTM[24] | 71.0 | 67.9 | 65.8 |
| RF-MediSys[25] | 72.3 | 69.7 | 67.2 |
| DITTO[26] | 74.6 | 72.8 | 70.4 |
| Ours | 77.8 | 75.9 | 73.2 |

The results indicate that the proposed method consistently outperforms baseline models, achieving 77.8% accuracy, with significant gains over LSTM and BiLSTM. It also achieves high precision (75.9%) and recall, demonstrating strong capability in identifying relevant diagnostic labels from unstructured EHR text. These improvements stem from the model's use of attention mechanisms and deep semantic representations, enabling better handling of context and label dependencies. Overall, the method offers a robust solution for multi-label disease prediction in clinical settings. The impact of learning rate on performance is further analyzed in Figure 2.

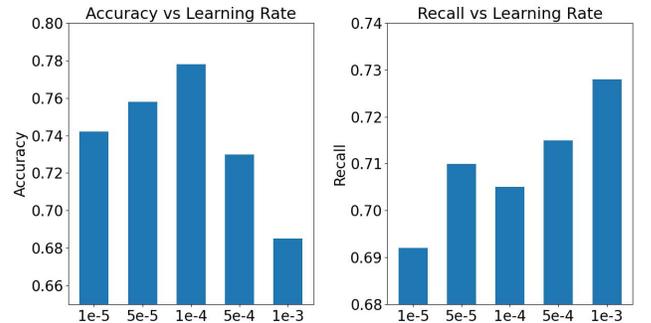

Figure 2. Analysis of the impact of different learning rate settings on model performance

The figure illustrates that learning rate plays a critical role in model performance for multi-label medical information extraction. Accuracy peaks at a moderate learning rate of 1e-4 (0.778), indicating optimal trade-offs between convergence and semantic modeling. Recall, however, increases steadily with higher learning rates, rising from 0.689 to 0.731 as the rate moves from 1e-5 to 1e-3, suggesting improved coverage but at the cost of reduced accuracy. This reflects a precision-recall trade-off common in medical prediction tasks, where maximizing label coverage risks introducing noise. The results highlight the need to carefully tune learning rates to balance predictive accuracy and recall. A further sensitivity analysis on varying training sample sizes is presented in Figure 3.

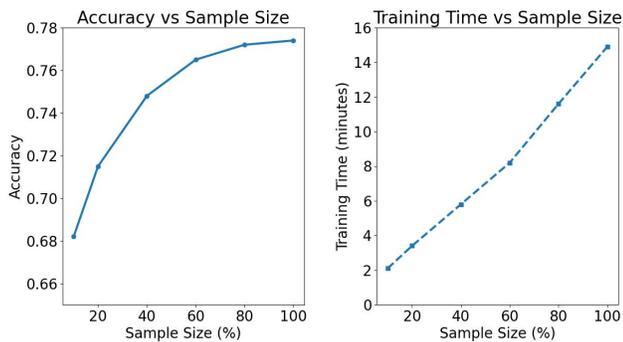

Figure 3. Sensitivity evaluation of different sample size ratios on training effects

The results in the figure show that sample size proportion has a significant impact on model performance. As the training sample increases from 10% to 100%, the model's accuracy improves steadily. The increase is particularly notable between 20% and 60%, suggesting that the model struggles to learn the latent patterns and semantic structures in medical records when data is limited. When the sample size exceeds 80%, the accuracy begins to stabilize. This indicates that the model has captured most of the essential features required for the multi-label disease prediction task.

This trend in accuracy reflects the data-driven nature of modeling electronic medical records. The scale of data directly determines the model's ability to capture associations between diagnostic labels and contextual information. In complex and semantically rich datasets like MIMIC-IV, insufficient data often limits the model's ability to recognize rare diseases or weakly expressed labels. As the sample size grows, the model gains better coverage of diverse clinical scenarios and improves its semantic understanding both broadly and deeply.

However, training time, as another sensitivity factor, also increases linearly with sample size. When using 100% of the data, the training time reaches nearly 15 minutes, which is over seven times that of using only 10% of the data. This suggests that while performance gains are desirable, attention must also be paid to computational cost and time consumption. In real-world clinical deployments, model inference efficiency and scalability are critical factors in assessing practical value.

This paper further conducts an experiment to investigate the effect of introducing noise into the input text on the robustness of the model. By deliberately injecting different forms of perturbations into the textual data, the study aims to simulate real-world scenarios where input quality may vary. This allows for a comprehensive evaluation of the model's stability and its ability to maintain performance under less-than-ideal conditions. The experimental results derived from this analysis are presented in Figure 4.

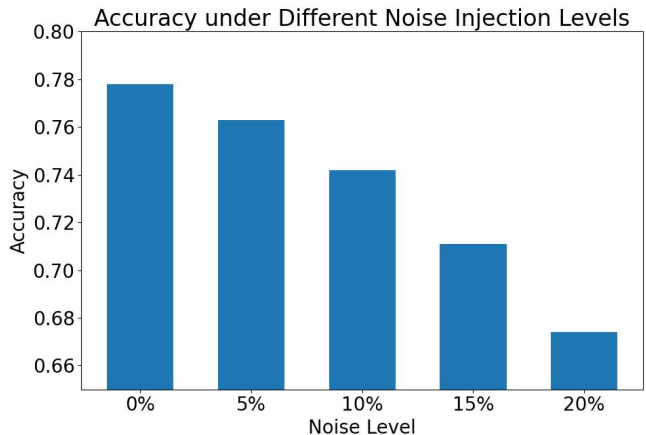

Figure 4. Experiment on the effect of noise injection text on model robustness

The figure reveals that increasing noise levels lead to a gradual decline in model accuracy, highlighting sensitivity to input perturbations in multi-label disease prediction. Accuracy remains relatively stable up to 10% noise but drops sharply below 0.67 at 20%, indicating limited robustness to severe semantic degradation. While the model demonstrates moderate fault tolerance—likely due to its ability to capture contextual dependencies—it struggles when critical medical terms are disrupted. This underscores the importance of semantic integrity in EHR-based tasks. Although self-attention aids structural modeling, performance remains vulnerable to textual noise. These findings suggest that enhancing robustness—through techniques like semantic augmentation or contrastive learning—will be essential for improving model stability in real-world clinical environments.

## V. CONCLUSION

This study focuses on information extraction and multi-label disease prediction from electronic health record texts. A deep modeling method based on attention mechanisms is proposed. The method is designed to address the structural complexity of medical language and the diversity of label distribution. By introducing the Transformer architecture, the model can retain rich contextual semantic information while capturing potential dependencies among labels. This enables effective representation and decoding of high-dimensional, unstructured medical text. Experimental results show that the proposed method achieves superior performance in accuracy, precision, and recall on real ICU patient records. It demonstrates strong potential for practical application.

In terms of model design, the method integrates the characteristics of multi-label tasks with the semantic properties of medical text. A multi-layer attention mechanism is used to dynamically focus on key information, significantly improving the recognition of co-occurring diseases and weak labels. Meanwhile, in a series of sensitivity analyses, the method maintains strong robustness under various real-world conditions, including sample size variation, parameter perturbation, and input noise. This stability is critical for clinical decision support and automated health management systems. It helps reduce human error and enhances the efficiency of diagnosis and treatment.

This study contributes not only a generalized and scalable model framework but also a systematic exploration of medical natural language processing applications. Through empirical analysis, it clarifies the relationship between data volume, model capacity, and noise tolerance. These findings provide a foundation for building more stable and fine-grained intelligent medical systems. In addition, the method shows strong adaptability across datasets, offering a reusable technical solution for information extraction tasks across different scenarios and disease types. It holds significant potential for real-world deployment.

## VI. Future work

Future work may proceed in two directions. On the modeling side, incorporating graph-based knowledge, time series modeling, and multimodal perception can further enhance semantic understanding in complex medical contexts. On the application side, introducing active learning and human-in-the-loop mechanisms may improve the model's adaptability in data-sparse and imbalanced scenarios. With the ongoing accumulation of medical big data and the improvement of health information systems, this research offers theoretical foundations and engineering insights for building intelligent and precise health information processing platforms.